\documentclass[conference,dvipsnames]{IEEEtran}

\usepackage{balance}
\usepackage{booktabs}
\usepackage{todonotes}
\usepackage{bm}
\usepackage{amsmath}
\usepackage{comment}
\usepackage[algo2e, ruled, linesnumbered]{algorithm2e}
\usepackage{subfigure}
\usepackage{url}
\usepackage[hidelinks]{hyperref}

\let\svthefootnote\thefootnote
\newcommand\freefootnote[1]{%
  \let\thefootnote\relax%
  \footnotetext{#1}%
  \let\thefootnote\svthefootnote%
}

\begin{document}

\title{Dying Clusters Is All You Need - Deep Clustering With an Unknown Number of Clusters}
\makeatletter
\newcommand{\linebreakand}{%
  \end{@IEEEauthorhalign}
  \hfill\mbox{}\par
  \mbox{}\hfill\begin{@IEEEauthorhalign}
}
\makeatother

\author{
\IEEEauthorblockN{Collin Leiber$^*$, Niklas Strauß$^*$, Matthias Schubert, Thomas Seidl}
\IEEEauthorblockA{
\textit{Munich Center for Machine Learning (MCML), LMU Munich} \\
Munich, Germany \\
\{leiber, strauss, schubert, seidl\}@dbs.ifi.lmu.de
}
}

\maketitle
\freefootnote{$^*$Authors contributed equally.}

\begin{abstract}
Finding meaningful groups, i.e., clusters, in high-dimensional data such as images or texts without labeled data at hand is an important challenge in data mining. In recent years, deep clustering methods have achieved remarkable results in these tasks. However, most of these methods require the user to specify the number of clusters in advance. This is a major limitation since the number of clusters is typically unknown if labeled data is unavailable. Thus, an area of research has emerged that addresses this problem. Most of these approaches estimate the number of clusters separated from the clustering process. This results in a strong dependency of the clustering result on the quality of the initial embedding. Other approaches are tailored to specific clustering processes, making them hard to adapt to other scenarios. In this paper, we propose UNSEEN, a general framework that, starting from a given upper bound, is able to estimate the number of clusters. To the best of our knowledge, it is the first method that can be easily combined with various deep clustering algorithms. We demonstrate the applicability of our approach by combining UNSEEN with the popular deep clustering algorithms DCN, DEC, and DKM and verify its effectiveness through an extensive experimental evaluation on several image and tabular datasets. Moreover, we perform numerous ablations to analyze our approach and show the importance of its components. The code is available at: \url{https://github.com/collinleiber/UNSEEN}
\end{abstract}


\section{Introduction}
Identifying clusters in data without supervision is an important task in machine learning. As a result, clustering is a well-studied area of research, and many different approaches have been proposed~\cite{clusteringSurvey}. Since traditional methods struggle with high-dimensional data, like images and texts, many state-of-the-art clustering algorithms are based on deep learning. This enables them to learn powerful representations to effectively cluster the data without using label information.

A large group of these so-called deep clustering algorithms applies autoencoders (AEs)~\cite{autoencoder} to learn low-dimensional representations in which to cluster the data effectively~\cite{deepClusteringSurvey}.
AE-based clustering algorithms are popular because they not only work well with image data but have been successfully applied to a vast array of different data domains such as text or tabular data~\cite{dec}. 
In contrast to more sophisticated strategies, like contrastive networks, AEs are often easier to apply as they do not rely on augmentations, which can be difficult to define for certain data types such as tabular data. Consequently, AE-based deep clustering procedures can be applied to a broad range of different types of data.
Due to their versatility, we focus on AE-based deep clustering algorithms in the following.
Even though these approaches work on unlabeled data, the vast majority of these algorithms still require the user to specify the number of clusters in advance. However, the number of clusters is usually unknown and particularly challenging to obtain due to the lack of label information/ground truth.

To overcome this issue, a number of procedures have been developed for classical (non-deep) clustering algorithms~\cite{stopElbow}.
However, these approaches can be ineffective when applied to high-dimensional data as they struggle with the curse of dimensionality~\cite{curseOfDimensionality}. Furthermore, they often rely on unsupervised metrics, such as the silhouette score~\cite{silhouetteScore}, which can become meaningless in a deep learning environment as the data spaces can change significantly in each iteration.

Thus, they are often unsuitable for use in combination with deep clustering algorithms. 
Some approaches to estimate the number of clusters have been proposed specifically for deep clustering algorithms. These approaches can be divided into two main categories. The former work in a sequential manner, i.e., they first learn an embedding, then estimate the number of clusters based on the embedding, and cluster the data~\cite{scde, ddc}. However, these approaches are heavily dependent on the initial embedding and can, therefore, be quite unstable. The latter propose specialized loss functions tailored to specific algorithms~\cite{dipdeck, deepdpm} that are highly specialized and cannot be easily applied to other clustering algorithms. 

To the best of our knowledge, currently no general framework exists to estimate the number of clusters that can be straightforwardly applied to a multitude of deep clustering algorithms. In this paper, we propose UNSEEN (\textbf{u}nknown \textbf{n}umber of clu\textbf{s}ters in d\textbf{ee}p embedi\textbf{n}gs), a general framework that can be applied to many deep clustering algorithms -- including prominent examples such as DCN~\cite{dcn}, DEC~\cite{dec}, or DKM~\cite{dkm} -- without specifying the exact number of clusters. Our approach merely requires to specify an upper bound on the number of clusters, which can be more easily obtained. Furthermore, we empirically demonstrate that our approach achieves good performances, even with upper bounds that vastly overestimate the actual number of clusters. Thus, we argue that it is not particularly difficult to provide such a loose upper bound in most applications.

\section{Related Work}
This work addresses the question of how to find high-quality clusters in deep clustering if the exact number of clusters is unknown. Therefore, two main branches of research are relevant: deep clustering and estimating the number of clusters.

\textbf{Deep Clustering:} We focus on AE-based deep clustering approaches, as they are not restricted to certain data types. This perfectly matches our goal of designing a generic framework that can be applied to a majority of algorithms and datasets.

Deep clustering procedures based on AEs can be divided into three main categories: sequential, alternating, and simultaneous approaches~\cite{deepClusteringSurvey}. In sequential approaches, an embedding embedding is learned first and afterward the clustering task is performed on the resulting embedded data. A well-known approach is to train an AE using the reconstruction loss $\mathcal{L}_\text{rec}$ --- often the mean squared error is utilized --- and then execute $k$-Means~\cite{kmeansLloyd}. We denote this strategy as AE+$k$-Means. Alternating deep clustering algorithms go one step further by creating an embedding tailored to the clustering task at hand. Therefore, they perform a batch-wise optimization of a clustering-specific loss function $\mathcal{L}_\text{clust}$ together with the reconstruction loss. The combined loss for a given batch $\mathcal{B}$ can be defined as: 
\begin{equation*}
    \mathcal{L}_\text{total}(\mathcal{B}) = \lambda_1 \mathcal{L}_\text{rec}(\mathcal{B}) + \lambda_2 \mathcal{L}_\text{clust}(\mathcal{B}),
\end{equation*}
where $\lambda_1$ and $\lambda_2$ are parameters to balance the two loss terms.
Since alternating approaches often use hard cluster labels, their loss function is usually not fully differentiable. Therefore, they iterate between optimizing the embedding and optimizing the clustering result. 

A common way to define $\mathcal{L}_\text{clust}$ is using the squared Euclidean distance between the embedded samples and their associated cluster center, i.e.,
\begin{equation*}
    \mathcal{L}_\text{clust}(\mathcal{B}) = \sum_{i \in [1, k]}\sum_{x\in C_i}||\text{enc}(x)-\mu_i||^2_2,
\end{equation*}
where $\mu_i$ is the cluster center of cluster $i$. Prominent examples utilizing this loss function include AEC~\cite{aec} and DCN~\cite{dcn}. Other methods use, for example, ideas from subspace clustering~\cite{acedec} or statistical tests~\cite{dipencoder}.

In contrast to alternating deep clustering approaches, simultaneous algorithms optimize the embedding and the clustering parameters simultaneously by using soft cluster assignments. The most well-known representative is DEC~\cite{dec}, where $\mathcal{L}_\text{clust}$ is based on the Kullback-Leibler divergence to compare a centroid-based data distribution with an auxiliary target distribution. While DEC ignores $\mathcal{L}_\text{rec}$ during the clustering optimization, IDEC~\cite{idec} was proposed as an extension of DEC that uses $\mathcal{L}_\text{rec}$ also during the clustering optimization. DCEC~\cite{dcec} is a variant of IDEC that showcases that the applied feedforward AE can be easily substituted by a convolutional AE.
The algorithm DKM~\cite{dkm} proposes a different loss function which is more similar to the classic $k$-Means procedure. Here, the Euclidean loss is extended by a factor based on a parameterized softmax function to simulate hard cluster labels. Thereby, the embedding and the cluster centers are fully differentiable.

\textbf{Estimating the number of clusters:}
A prominent branch of algorithms that can estimate the number of clusters is density-based approaches. These procedures usually follow the idea of DBSCAN~\cite{dbscan} and search for areas in the feature space that contain neighborhoods of data points. While such approaches work well in many scenarios it is non-trivial to transfer these ideas into a deep learning setting. Furthermore, often complex parameters have to be set to define the necessary characteristics of a neighborhood.

Another line of research extends centroid-based clustering algorithms like $k$-Means~\cite{kmeansLloyd} so that the number of centers can be estimated. These approaches can be subdivided into top-down processes which start with a low number of clusters that is successively increased and bottom-up processes which work the opposite way. Examples for top-down approaches include the Gap Statistic~\cite{gapStatistic}, $X$-Means~\cite{xMeans}, $G$-Means~\cite{gMeans}, $PG$-Means~\cite{pgmeans}, and DipMeans~\cite{dipmeans}. 
An prominent bottom-up approach is proposed in~\cite{mdlBischof}.
In both scenarios, a common way to decide which clusters to split or merge is to utilize concepts from information theory or statistical tests like Anderson–Darling~\cite{andersonDarling} or the dip-test of unimodality~\cite{diptest}.

In this paper, we integrate ideas from density-based clustering into a novel bottom-up deep clustering approach by leveraging nearest-neighbors.

\textbf{Deep Clustering + Estimating the number of clusters:}
Although deep clustering algorithms that are able to estimate the number of clusters exist already, they exhibit severe disadvantages. A common paradigm is to define sequential deep clustering approaches that combine the transformation capabilities of a neural network with established or novel $k$-estimation techniques. This is done, for example, by SCDE~\cite{scde}, DDC~\cite{ddc}, and DeepDPM$_\text{two-step}$~\cite{deepdpm}. However, these algorithms heavily depend on the quality of the initial embedding as they can not fix certain errors at a later stage. Furthermore, it can be hard to decide beforehand which $k$-estimation technique best matches the resulting embedding.

Other approaches, like DipDECK~\cite{dipdeck} or DeepDPM$_\text{end-to-end}$~\cite{deepdpm}, propose specific loss functions which are hard to adapt to other cluster definitions. In contrast, our UNSEEN framework is defined in such a way that it can be easily combined with a vast array of deep clustering approaches.

\section{A General Framework for Deep Clustering With an Unknown Number of Clusters}

In this paper, we build upon the observation that samples oftentimes change their cluster assignments during training. This is because the learned embeddings of deep clustering algorithms are very flexible. We observed that this can lead to clusters becoming essentially empty, a phenomenon we refer to as dying clusters. However, even though these clusters only contain very few points, deep clustering algorithms typically keep these clusters because the number of clusters is fixed. Our approach leverages these dying clusters and removes them, resulting in a final clustering with a lower number of clusters. In the following, we will detail our approach.

First, we define a cluster $C_i$ as the set of samples assigned to this cluster.
We use the notation $C_i^j$ to refer to cluster $i$ in epoch $j$ and $|C_i^j|$ to denote its current size (i.e., the number of samples in this cluster). Additionally, we also define a creation size $|C_i^0|$ for each cluster. The creation size of a cluster is initialized to the size of the cluster before the first epoch. Importantly, the creation size of a cluster is independent of its current size and only updated when the cluster receives samples from a dead cluster.
When the size of a cluster $|C_i^j|$ shrinks below a fraction $t$ of its creation size $|C_i^0|$, we refer to the cluster as dead and remove it completely. That means, the samples corresponding to the dead cluster are reassigned to the remaining active clusters and the creation sizes of the active clusters are updated accordingly. More specifically, the creation size $|C_i^0|$ of the cluster is increased by the number of samples that are reassigned to it from dead clusters. In order to remove the dead clusters from the underlying deep clustering algorithm, the parameter $k$, which specifies the number of clusters, is updated to the current number of active clusters $k_j$. The samples of the dead clusters are reassigned to the most likely or closest cluster in accordance with the deep clustering algorithm.

After each epoch, we identify dead clusters and remove them. In contrast to most deep clustering algorithms, where the number of cluster remains constant, this leads to a decrease in the number of clusters over time.
Let us note that this process bears some resemblance to~\cite{dipdeck}. However, in contrast to~\cite{dipdeck} our approach is not tailored to a specific clustering algorithm. Instead, our method can be easily combined with various deep clustering algorithms.

Our approach converges to a fixed amount of clusters over time. However, this process and, thus, the final clusters are largely influenced by the initial clustering. In fact, many methods encounter difficulties in overcoming this initial bias, as they tend to strongly optimize in the direction of the initial clustering. This may result in a subpar final clustering performance. To counteract the impact of the initial bias on the clustering, we propose to add a nearest-neighbor-based loss term, which brings neighboring clusters closer together in the latent space. Additionally, we believe that this loss also pulls samples towards larger clusters, which can enhance the clustering quality in balanced datasets. Moreover, we observe that the loss acts as a regularization term that facilitates samples to change their cluster during training. This is because we only select the nearest neighbors from samples within the batch. Consequently, the nearest neighbors are likely to come from different clusters and, due to the Euclidean distance, are pulled closer to the other cluster. 
More formally, our nearest-neighbor loss is defined as follows:
\begin{equation*}
    \mathcal{L}_\text{UNSEEN}(\mathcal{B}, l) = \frac{1}{l|\mathcal{B}|} \sum_{x \in \mathcal{B}} \sum_{y\in \text{nn}(\mathcal{B}, x, l)} ||\text{enc}(x)-\text{enc}(y)||_2^2,
\end{equation*}
where $\mathcal{B}$ corresponds to a batch of data, $\text{enc}(x)$ is the embedded version of $x$, $\text{nn}(\mathcal{B}, x, l)$ returns the $l$-nearest-neighbors of $\text{enc}(x)$ in $\{\text{enc}(y) \mid y \in \mathcal{B}\}$ and $||\cdot||_2^2$ denotes the squared Euclidean distance.
Our empirical findings show that this nearest-neighbor loss is effective for sequential deep clustering approaches like DCN~\cite{dcn}. However, when combined with simultaneous methods, such as DEC~\cite{dec} or DKM~\cite{dkm}, we observe that the loss can cause the embeddings to collapse. In these algorithms, the clustering parameters are updated together with the embedding, which can result in significant changes to the clustering structure and, as a consequence, the formation of excessively large individual clusters. To inhibit the formation of such oversized individual clusters, we normalize the nearest-neighbor loss by considering all pairwise distances in a batch. This reduces the impact of the change in embeddings. 
Thus, when dealing with a simultaneous deep clustering algorithm, we use the following modified nearest-neighbor loss:
\begin{equation*}
    \mathcal{L}_\text{UNSEEN}^\text{simul}(\mathcal{B}, l) = \frac{\mathcal{L}_\text{UNSEEN}(\mathcal{B}, l)}{\frac{1}{|\mathcal{B}|(|\mathcal{B}| - 1)} \sum_{x \in \mathcal{B}} \sum_{\substack{y \in \mathcal{B}\\x \neq y}} ||\text{enc}(x)-\text{enc}(y)||_2^2}  
\end{equation*}
This results in the complete loss function
\begin{equation}
    \label{eq:totalLoss}
    \mathcal{L}_\text{total}(\mathcal{B}, l) = \lambda_1 \mathcal{L}_\text{rec}(\mathcal{B}) + \lambda_2 \left(\mathcal{L}_\text{clust}(\mathcal{B}, C) + \mathcal{L}_\text{UNSEEN}(\mathcal{B}, l)\right),
\end{equation}
where $\mathcal{L}_\text{rec}(\mathcal{B})$ denotes the reconstruction loss of the AE, $\mathcal{L}_\text{clust}(\mathcal{B}, C)$ the clustering loss of the underlying deep clustering algorithm, and $\lambda_1$ as well as $\lambda_2$ are weighting terms.

\begin{algorithm2e}[t]
	\SetAlgoVlined
	\DontPrintSemicolon
	\KwIn{dataset $\mathcal{X}$, initial number of clusters $k_\text{init}$, dying threshold $t$, number of epochs $e$}
	\KwOut{the final clustering model $M$}
    pretrained neural network $NN$\;
    initialize clustering model $M$ using $k_\text{init}$\;
    extract initial cluster assignments $C_i^0$ from $M$\;
    $j = 1$\;
	\While{$j \le e$}{
        compute $l$ (Eq. \ref{eq:nearestNeighbors})\;
	    \For{$\mathcal{B}$ \textbf{in} $\text{batches}(\mathcal{X})$}{
            calculate $\mathcal{L}_\text{UNSEEN}(\mathcal{B}, l)$ (Eq. \ref{eq:totalLoss})\;
            optimize $NN$ using $\mathcal{L}_\text{UNSEEN}$\;
	    }
        in case of an alternating base algorithm, update $M$\;
        // Check for dead clusters\;
        extract current cluster assignments $C_i^j$ from $M$\;
        identify dead clusters by checking $\frac{|C_i^j|}{|C_i^0|}<t$\;
        \If{dead clusters exist}{
            remove dead clusters form $M$\;
            extract new cluster assignments $C_i^j$ from $M$\;
            update $C_i^0$\;
        }
        $j = j+1$\;
	}
	\Return{$M$}
	\caption{Process of UNSEEN}
 \label{alg:UNSEEN}
\end{algorithm2e}

Our nearest-neighbors loss term has introduced an additional hyperparameter. Namely, the number of nearest neighbors $l$ to consider. At first glance, selecting a suitable value for this parameter appears to be challenging. Especially, since a good value depends on the number of currently active clusters, which varies during the optimization process. However, we propose to set this parameter based on the following intuition. In the beginning, samples of different clusters are usually still close in the embedding; thus, a large number of nearest neighbors would lead to pulling samples from different clusters together, deteriorating the effectiveness of the clustering. However, later in the clustering process, when many clusters have already died, they are usually more compact and further separated in the latent space. Thus, we want to include more neighbors in order to focus more on the neighboring clusters and facilitate the combination of clusters by strengthening the connection between these clusters. Therefore, we set the number of nearest neighbors $l$ dependent on the number of currently active clusters. Formally, it is defined as follows:
\begin{equation}
    \label{eq:nearestNeighbors}
    l=\frac{|\mathcal{B}|}{k_j}.
\end{equation}
Here, $|B|$ denotes the batch size and $k_j$ the number of active clusters in epoch $j$.

We detail the complete process of our framework in Algorithm~\ref{alg:UNSEEN}. In contrast to existing approaches, it is straightforward to integrate our method into most deep clustering algorithms. More specifically, our framework consists of two components. The first component is an additional loss term, which is trivial to integrate into deep clustering algorithms. The second part of our framework keeps track of active clusters and dissolves dead clusters. This is done outside of the underlying clustering optimization and thereby independent of it.

\section{Experiments}

We thoroughly evaluate the effectiveness of our framework in various synthetic and real-world benchmarks by comparing its performance against popular deep clustering baselines, including commonly used sequential and simultaneous approaches. Furthermore, we provide a detailed analysis of UNSEEN's behavior and properties. Finally, we perform extensive ablations to empirically examine our design choices and parameters. More specifically, we focus on the following research questions:
\begin{itemize}
    \item \textbf{RQ1}: How good are the clustering results of UNSEEN in combination with different deep clustering algorithms and how do they compare to the baselines across different datasets?
    \item \textbf{RQ2}: How accurately can UNSEEN estimate the number of clusters?
    \item \textbf{RQ3}: What is the impact of the nearest-neighbor loss $\mathcal{L}_\text{UNSEEN}$ on the performance of UNSEEN?
    \item \textbf{RQ4}: How sensitive is UNSEEN to the value of the dying threshold $t$?
\end{itemize}

\textbf{Datasets:}
We perform our experiments on various image datasets, including Optdigits~\cite{optdigits}, USPS~\cite{usps}, MNIST~\cite{mnist}, Fashion-MNIST (FMNIST)~\cite{fmnist}, and Kuzushiji-MNIST (KNNIST)~\cite{kmnist}.
Additionally, we use the tabular dataset Pendigits~\cite{pendigits}.
Further details regarding the datasets are shown in the first column of Table~\ref{tab:clusteringResults}. We pre-process the data in accordance with~\cite{dipdeck}. This includes a channel-wise z-normalization ($0$ mean and standard deviation of $1$) and flattening for image datasets and a feature-wise z-normalization for tabular datasets.

\textbf{Evaluation metrics:}
To evaluate the clustering results, we use the following three commonly used metrics: Normalized Mutual Information (NMI)~\cite{nmi}, Adjusted Rand Index (ARI)~\cite{ari}, and Unsupervised Clustering Accuracy (ACC)~\cite{acc}. These metrics compare the predicted labels with a given ground truth. Let us note that a value of $1$ equals a perfect clustering, and lower values indicate that the clustering result is more similar to that of random clustering. 

\textbf{Experimental setting:}
We verify the broad applicability of our framework by combining UNSEEN with the alternating approach DCN~\cite{dcn} as well as the simultaneous approaches DEC~\cite{dcn} and DKM~\cite{dcn}, where DKM uses the reconstruction loss also during the clustering process and DEC does not.
To ensure comparability and a sound evaluation, our experimental setting follows the standard described in~\cite{deepClusteringBenchmark}. We use an autoencoder (AE) with the encoder layers of size $d$-$500$-$500$-$2000$-$10$. Here, $d$ denotes the number of input features in the original dataset. The decoder is a mirrored version of the encoder layers. This corresponds to the architecture used in many publications like~\cite{dec},~\cite{idec},~\cite{dcn},~\cite{dipencoder}, and~\cite{dkm}. All AEs are pre-trained for $100$ epochs using the mean squared error with a learning rate of $1e{-3}$, and we use the same pre-trained AEs as the basis for all algorithms. Afterward, the clustering process is executed for $150$ epochs with a learning rate of $1e{-4}$. In all experiments, ADAM~\cite{adam} is used as the optimizer with a batch size of $256$. For DKM and UNSEEN+DKM, we set the weights of the reconstruction loss $\lambda_1$ and clustering loss $\lambda_2$ to $1$ and the $\alpha$ parameter to $1000$. For DCN and UNSEEN+DCN, we set the weight of the reconstruction loss $\lambda_1$ to $1$ and the clustering loss $\lambda_2$ to $0.05$. As DEC does not use $\mathcal{L}_\text{rec}$ during clustering optimization, we do not have to set weights. To set the initial number of clusters, we follow the proposal in~\cite{dipdeck} and choose a value of $k_\text{init}=35$. Furthermore, we use a default dying threshold of $t=0.5$. The baselines DKM, DEC, and DCN are given the correct number of ground truth clusters. All implementations are based on ClustPy~\cite{deepClusteringBenchmark}. 

\subsection{RQ1: Clustering Results}

\begin{table*}[t]
\centering
\caption{Clustering results of UNSEEN and the underlying deep clustering methods, which were given the true number of clusters. $N=$ number of samples, $d=$ number of features, $k=$ number of true clusters. Each experiment is repeated ten times, and the average result ($\pm$ standard deviation) is stated in \%. The best average result for each dataset and metric is highlighted in \textbf{bold}, and the second-best is \underline{underlined}. An asterisk * indicates that UNSEEN is able to outperform the corresponding baseline algorithm.\label{tab:clusteringResults}}
\resizebox{1\textwidth}{!}{
\begin{tabular}{l|l|ccccccc}
\toprule
\textbf{Dataset} & \textbf{Metric} & AE+$k$-Means & UNSEEN+DCN & UNSEEN+DEC & UNSEEN+DKM & DCN & DEC & DKM\\
\midrule
Optdigits~\cite{optdigits} & ACC & $84.3 \pm 5.4$ & $81.3 \pm 6.2$ & $81.8 \pm 6.8$ & \underline{$85.6 \pm 3.3~*$} & $83.7 \pm 4.3$ & \bm{$86.0 \pm 3.9$} & $84.3 \pm 4.2$\\
{\tiny ($N=5620, d=64, k=10$)}& ARI & $74.8 \pm 5.6$ & $79.6 \pm 4.2~*$ & \underline{$80.6 \pm 4.8~*$} & \bm{$83.4 \pm 2.3~*$} & $76.2 \pm 4.3$ & $80.1 \pm 4.0$ & $76.6 \pm 4.9$\\
& NMI & $79.4 \pm 2.9$ & $85.6 \pm 1.5~*$ & \underline{$86.4 \pm 1.8~*$} & \bm{$87.1 \pm 1.1~*$} & $82.7 \pm 2.1$ & $86.3 \pm 2.0$ & $84.6 \pm 2.1$\\
\midrule
USPS~\cite{usps} & ACC & $68.3 \pm 2.9$ & \underline{$77.5 \pm 5.6~*$} & \bm{$81.3 \pm 2.4~*$} & $71.5 \pm 9.4~*$ & $69.2 \pm 5.7$ & $75.1 \pm 5.0$ & $58.1 \pm 4.7$\\
{\tiny ($N=9298, d=256, k=10$)} & ARI & $55.0 \pm 2.1$ & $69.8 \pm 4.9~*$ & \bm{$75.9 \pm 2.4~*$} & $62.6 \pm 10.6~*$ & $61.1 \pm 6.5$ & \underline{$70.2 \pm 4.5$} & $45.7 \pm 7.4$\\
& NMI & $65.5 \pm 1.6$ & $76.5 \pm 1.4~*$ & \bm{$80.8 \pm 0.9~*$} & $73.3 \pm 5.3~*$ & $71.9 \pm 2.9$ & \underline{$78.6 \pm 2.4$} & $62.9 \pm 3.8$\\
\midrule
MNIST~\cite{mnist} & ACC & $72.2 \pm 5.4$ & \bm{$86.0 \pm 5.4~*$} & $71.0 \pm 4.4$ & \underline{$83.0 \pm 5.5~*$} & $80.3 \pm 5.6$ & $77.2 \pm 6.4$ & $75.5 \pm 6.9$\\
{\tiny ($N=70000, d=784, k=10$)} & ARI & $61.8 \pm 5.1$ & \bm{$83.7 \pm 4.5~*$} & $70.9 \pm 3.5$ & \underline{$78.5 \pm 5.2~*$} & $75.9 \pm 4.4$ & $73.4 \pm 5.6$ & $67.3 \pm 7.4$\\
& NMI & $70.1 \pm 3.0$ & \bm{$87.8 \pm 1.9~*$} & $80.4 \pm 1.8$ & \underline{$84.2 \pm 2.6~*$} & $83.3 \pm 1.7$ & $81.8 \pm 2.9$ & $79.0 \pm 3.3$\\
\midrule
FMNIST~\cite{fmnist} & ACC & $56.4 \pm 6.6$ & \bm{$57.3 \pm 2.4~*$} & $54.4 \pm 3.8$ & $53.1 \pm 2.0$ & $56.8 \pm 3.9$ & \underline{$57.0 \pm 2.8$} & $54.7 \pm 3.2$\\
{\tiny ($N=70000, d=784, k=10$)} & ARI & $42.7 \pm 3.8$ & \bm{$46.9 \pm 1.5~*$} & $42.3 \pm 3.4$ & $43.9 \pm 1.4~*$ & \underline{$45.4 \pm 2.7$} & $43.3 \pm 3.0$ & $43.1 \pm 2.1$\\
& NMI & $56.7 \pm 2.2$ & \bm{$63.7 \pm 1.0~*$} & $58.6 \pm 2.6$ & \underline{$62.2 \pm 1.0~*$} & $61.3 \pm 2.1$ & $59.9 \pm 2.2$ & $59.9 \pm 2.0$\\
\midrule
KMNIST ~\cite{kmnist} & ACC & $55.6 \pm 2.6$ & $55.1 \pm 4.9$ & \underline{$59.2 \pm 1.8$} & $55.4 \pm 5.0~*$ & $55.8 \pm 3.9$ & \bm{$59.7 \pm 2.9$} & $46.8 \pm 5.7$\\
{\tiny ($N=70000, d=784, k=10$)} & ARI & $34.4 \pm 2.1$ & $39.2 \pm 4.3~*$ & \bm{$50.3 \pm 1.5~*$} & $37.6 \pm 10.0~*$ & $36.4 \pm 4.1$ & \underline{$41.1 \pm 2.1$} & $24.5 \pm 4.3$\\
& NMI & $47.4 \pm 1.5$ & $54.1 \pm 2.6~*$ & \bm{$64.3 \pm 1.0~*$} & \underline{$57.1 \pm 5.9~*$} & $52.6 \pm 2.6$ & $55.0 \pm 1.6$ & $46.0 \pm 3.7$\\
\midrule
Pendigits~\cite{pendigits} & ACC & $68.4 \pm 4.5$ & $74.6 \pm 3.6~*$ & \bm{$81.2 \pm 2.8~*$} & \underline{$78.3 \pm 1.9~*$} & $71.6 \pm 4.2$ & $76.5 \pm 4.2$ & $71.8 \pm 4.5$\\
{\tiny ($N=10992, d=16, k=10$)} & ARI & $53.9 \pm 3.1$ & $65.9 \pm 3.9~*$ & \bm{$71.9 \pm 3.7~*$} & \underline{$67.9 \pm 1.7~*$} & $58.6 \pm 4.8$ & $65.7 \pm 5.0$ & $58.5 \pm 5.0$\\
& NMI & $67.0 \pm 1.5$ & $76.3 \pm 1.3~*$ & \bm{$80.8 \pm 2.1~*$} & \underline{$78.9 \pm 0.8~*$} & $72.1 \pm 1.8$ & $76.6 \pm 3.0$ & $72.9 \pm 2.1$\\
\bottomrule
\end{tabular}}
\end{table*}

We start our evaluation by comparing the clustering results of our approach and its baselines on various datasets in Table~\ref{tab:clusteringResults}. Here, we can observe that UNSEEN almost consistently outperforms the baselines across all datasets and metrics. It achieves the best result in $16$ out of $18$ scenarios. Only for Optdigits and KMNIST, it is the second best according to the ACC metric. These results demonstrate the effectiveness of UNSEEN. Impressively, in the vast majority of the cases, it is able to outperform the underlying deep clustering algorithm that has been trained given the ground truth number of clusters. It is noteworthy that USEEN+DKM beats its baseline in all but one and UNSEEN+DCN in all but two scenarios. Furthermore, we observe that the standard deviation when using UNSEEN is usually similar to the underlying base algorithm, demonstrating that UNSEEN consistently produces good clusterings.

Let us briefly note that UNSEEN introduces a slight computational overhead. To estimate the extent, we compare the runtime necessary to compute the results from Table~\ref{tab:clusteringResults} of the original algorithms to the versions extended by UNSEEN. The increase in runtime is $+7\%$ for DCN, $+53\%$ for DEC, and $+42\%$ for DKM. It is easy to see that the increase in runtime for the simultaneous approaches DEC and DKM is significantly larger than for DCN. This is mainly due to the generally slower execution of alternating deep clustering algorithms. All experiments were executed on a computer with an NVIDIA RTX A6000 and an Intel Xeon Gold 6326 CPU. 

\subsection{RQ2: Correctness of the Estimated Number of Clusters}

\begin{figure}[t!]
\centering
\includegraphics[width=0.42\textwidth]{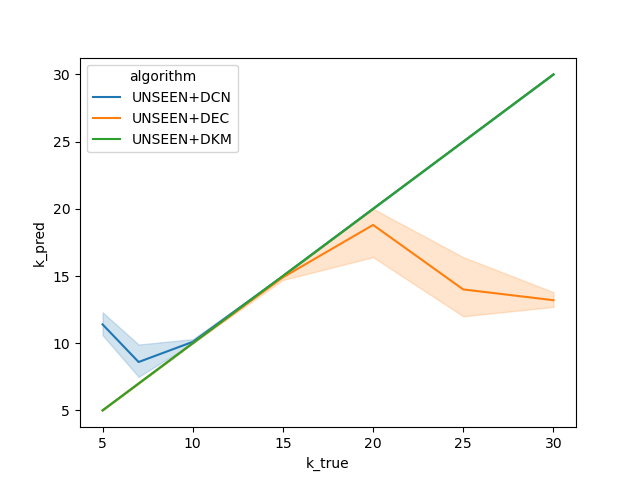}
\caption{Estimated number of clusters ($k\_pred$) for synthetic datasets with the true number of clusters ($k\_true$) within $[5, 30]$. The colored area marks the $95\%$ confidence interval.\label{fig:blobsK}}
\end{figure}

\begin{figure}[t!]
\centering
\includegraphics[width=0.42\textwidth]{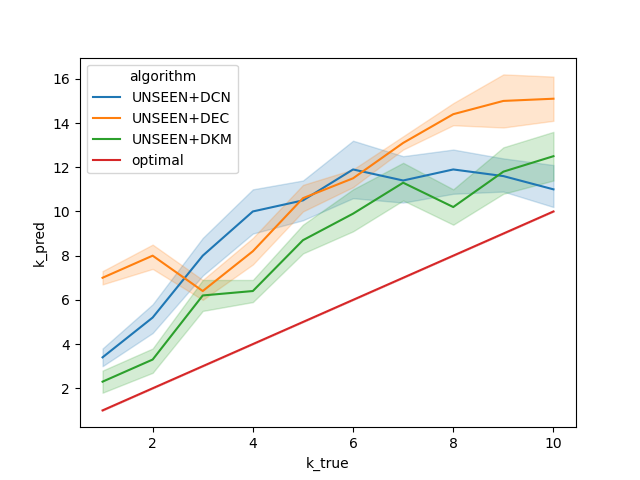}
\caption{Estimated number of clusters ($k\_pred$) for MNIST when considering only the first ($k\_true$) clusters. The colored area marks the $95\%$ confidence interval.\label{fig:mnistK}}
\end{figure}

We continue our evaluation by examining how well UNSEEN can estimate the number of clusters. For this, we use a synthetic and a real-world setting based on MNIST.

\textbf{Synthetic setting:} For the synthetic setting, we use the \textit{make blobs}\footnote{\url{https://scikit-learn.org/stable/modules/generated/sklearn.datasets.make_blobs.html}} functionality of scikit-learn~\cite{scikit-learn} to create datasets with a varying number of clusters $k \in \{5, 7, 10, 15, 20, 25, 30\}, N=10,000$ and $d=100$. Here, all clusters are made out of isotropic Gaussian blobs with a standard deviation of $1$ and contain an equal number of samples, i.e., $\forall_{1 \le j \le k}:|C_j|=\frac{N}{k}$.

The results are illustrated in Figure~\ref{fig:blobsK}. Note that the algorithms start with $k_\text{init}=35$ in all experiments and that this value marks the maximum number of clusters that can be reached. It can be observed that UNSEEN+DKM is able to correctly identify the number of clusters in all experiments and across all $7$ scenarios. UNSEEN+DCN overestimates the number of clusters in the beginning but correctly estimates the number of clusters for $k \ge 10$. For UNSEEN+DEC, it is the other way around. Here, the number of clusters is perfectly estimated for $k \le 15$ and underestimated thereafter. This may be because DEC does not take the reconstruction loss into account during cluster optimization, which allows greater deformations of the embedding.

\textbf{Real-world setting:} For the real-world setting, we take subsets of the MNIST dataset by only considering the first $k$ clusters, i.e., for $k=3$, we reduce MNIST to only contain the cluster labels $0$, $1$, and $2$. The results are shown in Figure~\ref{fig:mnistK}. All variants of UNSEEN overestimate the number of clusters as all results lie atop the red line, which indicates a perfect estimation. However, it is also notable that the number of estimated clusters increases with the number of ground truth clusters which corresponds to the desired behavior. The additional clusters could be due to different writing styles of the digits. Other studies have experienced a similar behavior~\cite{dipdeck, deepect, ddc}.
We would also like to emphasize that UNSEEN+DEC deviates the most from the true number of clusters, which may explain the slight decrease in performance between UNSEEN+DEC and DEC regarding MNIST as displayed in Table~\ref{tab:clusteringResults}.

\subsection{RQ3: Importance of $\mathcal{L}_\text{UNSEEN}$}

One of the core components of UNSEEN is the nearest-neighbor loss $\mathcal{L}_\text{UNSEEN}$. We perform an ablation study to empirically demonstrate the importance of this loss, which can be found in Table~\ref{tab:ignoringNNLoss}. While the results without $\mathcal{L}_\text{UNSEEN}$ are slightly better in some settings, this loss is crucial for the effectiveness of UNSEEN in other settings. Note that the maximum performance loss when using $\mathcal{L}_\text{UNSEEN}$ is $8.8\%$ in the case of the NMI result of UNSEEN+DCN for KMNIST. All other losses are markedly lower. The gains of applying $\mathcal{L}_\text{UNSEEN}$, however, are up to $36.0\%$ in the case of ACC for UNSEEN+DCN and Optdigts. Overall, we recommend including $\mathcal{L}_\text{UNSEEN}$ since it drastically improves the performance of UNSEEN on some datasets while it only slightly decreases the performance on others. 

To further investigate the influence of $\mathcal{L}_\text{UNSEEN}$ on the optimization, we visualize clustering results of UNSEEN+DEC with and without $\mathcal{L}_\text{UNSEEN}$ applied on USPS. For this task, we transform the ten-dimensional embedding into a two-dimensional representation using t-SNE~\cite{tsne}. The resulting data distributions are displayed in Figure~\ref{fig:tsne}. While the results after the first epoch look rather similar, it can be seen that UNSEEN+DEC without $\mathcal{L}_\text{UNSEEN}$ compresses the initial clusters more strongly. This separates them more vigorously from neighboring larger clusters and results in fewer dead clusters. Accordingly, we observe more clusters after epoch $100$.

\begin{table}[t]
\centering
\caption{Clustering results of our proposal when $\mathcal{L}_\text{UNSEEN}$ is ignored during the optimization. Notation corresponds to Table \ref{tab:clusteringResults}. The difference compared to the original experiment is given in brackets, where better results are highlighted in \textbf{bold}. \label{tab:ignoringNNLoss}}
\resizebox{0.5\textwidth}{!}{
\begin{tabular}{l|l|ccc}
\toprule
\textbf{Dataset} & \textbf{Metric} & UNSEEN+DCN & UNSEEN+DEC & UNSEEN+DKM\\
\midrule
Optdigits & ACC & $45.3 \pm 2.7~(-36.0)$ & $62.3 \pm 3.7~(-19.5)$ & $80.2 \pm 3.7~(-5.4)$\\
& ARI & $51.2 \pm 2.8~(-28.4)$ & $66.8 \pm 2.9~(-13.8)$ & $80.0 \pm 3.1~(-3.4)$\\
& NMI & $76.0 \pm 0.8~(-9.6)$ & $81.9 \pm 1.1~(-4.5)$ & $85.8 \pm 1.4~(-1.3)$\\
\midrule
USPS & ACC & $62.8 \pm 4.8~(-14.7)$ & $73.3 \pm 3.4~(-8.0)$ & \bm{$73.1 \pm 7.5~(+1.6)$}\\
& ARI & $61.2 \pm 3.3~(-8.6)$ & $71.6 \pm 2.9~(-4.3)$ & \bm{$62.9 \pm 9.4~(+0.3)$}\\
& NMI & $74.1 \pm 1.1~(-2.4)$ & $79.7 \pm 1.0~(-1.1)$ & \bm{$74.0 \pm 3.8~(+0.7)$}\\
\midrule
MNIST & ACC & $52.0 \pm 5.0~(-34.0)$ & $58.0 \pm 3.2~(-13.0)$ & $75.1 \pm 3.3~(-7.9)$\\
& ARI & $58.0 \pm 4.7~(-25.7)$ & $61.8 \pm 3.0~(-9.1)$ & $70.5 \pm 3.6~(-8.0)$\\
& NMI & $77.9 \pm 1.1~(-9.9)$ & $78.7 \pm 1.3~(-1.7)$ & $80.7 \pm 1.7~(-3.5)$\\
\midrule
FMNIST & ACC & $40.3 \pm 1.8~(-17.0)$ & $53.7 \pm 1.2~(-0.7)$ & $49.4 \pm 1.8~(-3.7)$\\
& ARI & $36.2 \pm 1.5~(-10.7)$ & \bm{$44.7 \pm 1.2~(+2.4)$} & $41.7 \pm 1.0~(-2.2)$\\
& NMI & $59.4 \pm 0.7~(-4.3)$ & \bm{$61.6 \pm 1.0~(+3.0)$} & $61.6 \pm 0.6~(-0.6)$\\
\midrule
KMNIST & ACC & $47.4 \pm 2.7~(-7.7)$ & $53.0 \pm 1.3~(-6.2)$ & $53.0 \pm 5.8~(-2.4)$\\
& ARI & \bm{$42.2 \pm 2.3~(+3.0)$} & $48.3 \pm 1.1~(-2.0)$ & $35.8 \pm 11.0~(-1.8)$\\
& NMI & \bm{$62.9 \pm 1.1~(+8.8)$} & \bm{$65.7 \pm 0.6~(+1.4)$} & \bm{$57.2 \pm 5.1~(+0.1)$}\\
\midrule
Pendigits & ACC & $50.5 \pm 2.4~(-24.1)$ & $69.6 \pm 2.8~(-11.6)$ & $71.1 \pm 4.3~(-7.2)$\\
& ARI & $50.7 \pm 2.0~(-15.2)$ & $68.6 \pm 2.7~(-3.3)$ & $66.7 \pm 3.2~(-1.2)$\\
& NMI & $74.0 \pm 0.8~(-2.3)$ & $80.6 \pm 1.1~(-0.2)$ & $78.7 \pm 1.3~(-0.2)$\\
\bottomrule
\end{tabular}}
\end{table}

\begin{figure*}[t]
	\centering
	\subfigure[With $\mathcal{L}_\text{UNSEEN}$ - Epoch 1.]{
		\includegraphics[width=0.235\textwidth]{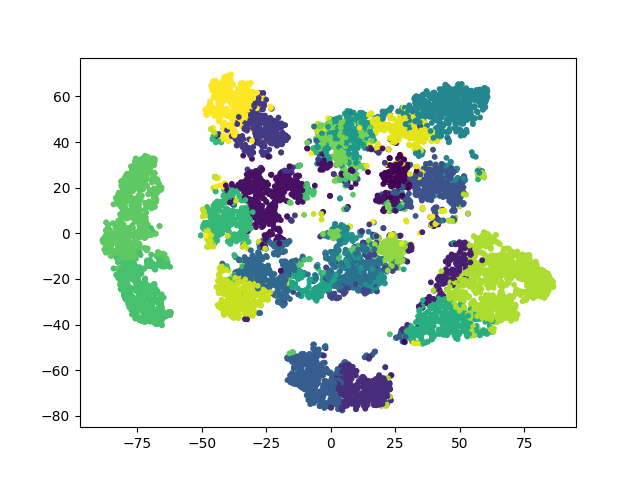}}
	\subfigure[With $\mathcal{L}_\text{UNSEEN}$ - Epoch 3.]{
	    \includegraphics[width=0.235\textwidth]{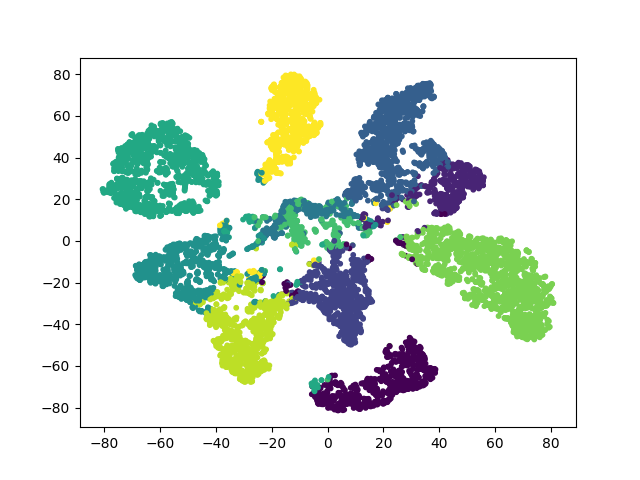}}
	\subfigure[With $\mathcal{L}_\text{UNSEEN}$ - Epoch 6.]{
		\includegraphics[width=0.235\textwidth]{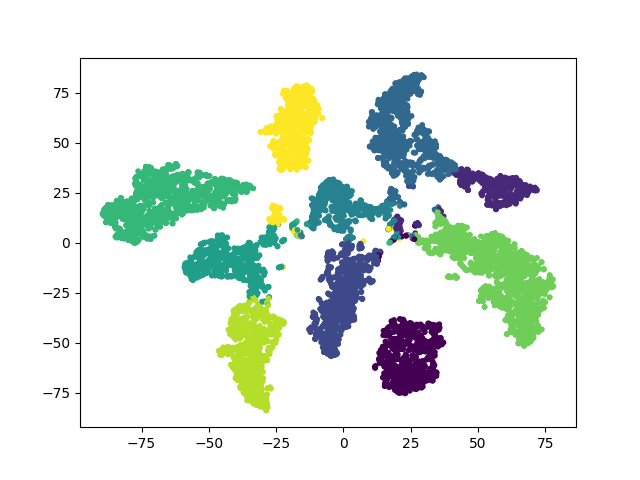}}
	\subfigure[With $\mathcal{L}_\text{UNSEEN}$ - Epoch 100.]{
		\includegraphics[width=0.235\textwidth]{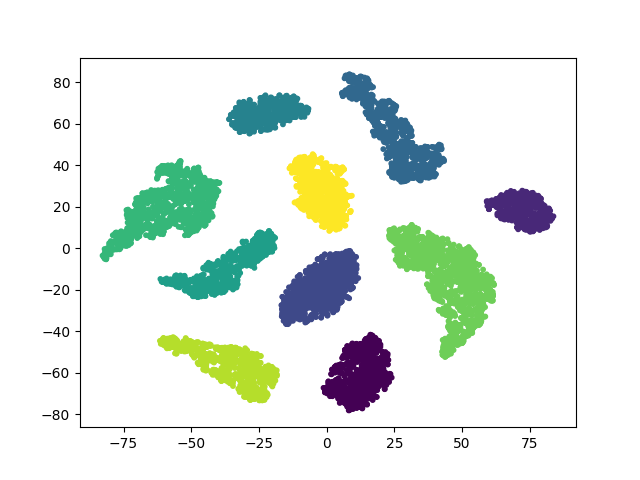}}
  	\subfigure[Without $\mathcal{L}_\text{UNSEEN}$ - Epoch 1.]{
		\includegraphics[width=0.235\textwidth]{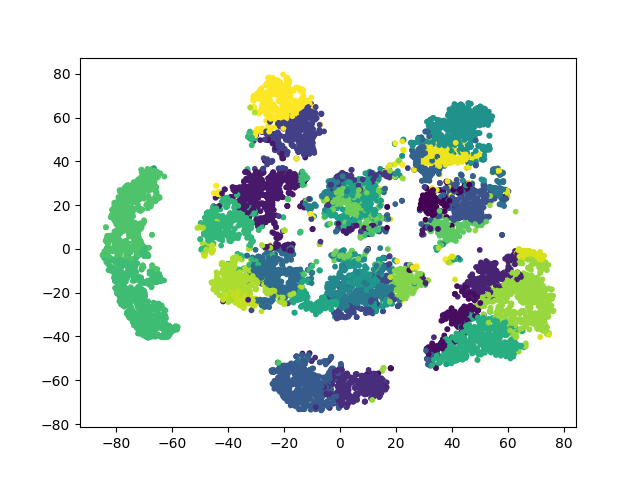}}
	\subfigure[Without $\mathcal{L}_\text{UNSEEN}$ - Epoch 3.]{
	    \includegraphics[width=0.235\textwidth]{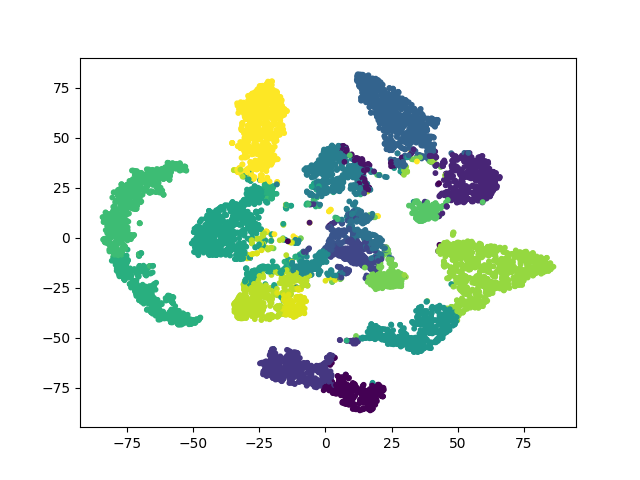}}
	\subfigure[Without $\mathcal{L}_\text{UNSEEN}$ - Epoch 6.]{
		\includegraphics[width=0.235\textwidth]{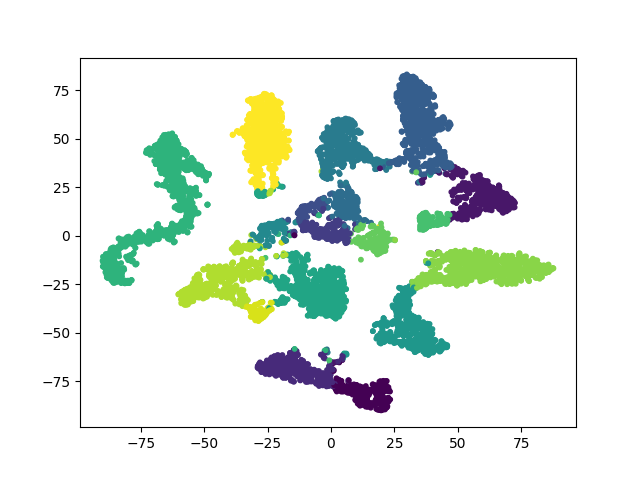}}
	\subfigure[Without $\mathcal{L}_\text{UNSEEN}$ - Epoch 100.]{
		\includegraphics[width=0.235\textwidth]{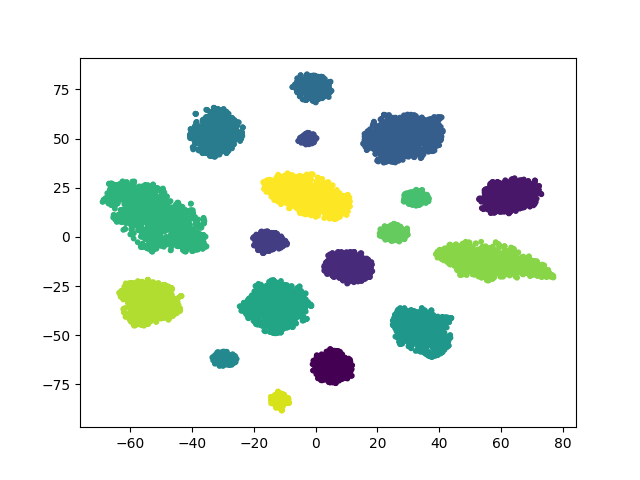}}
	\caption{Visualizations of different epochs of UNSEEN+DEC executed on USPS with and without using $\mathcal{L}_\text{UNSEEN}$. A two-dimensional representation of the embedding is obtained by applying t-SNE. Colors correspond to the current cluster labels.}
	\label{fig:tsne}
\end{figure*}

\subsection{RQ4: Impact of Dying Threshold $t$}

\begin{figure*}[h!]
	\centering
	\subfigure[Optdigits.]{\includegraphics[width=0.37\textwidth]{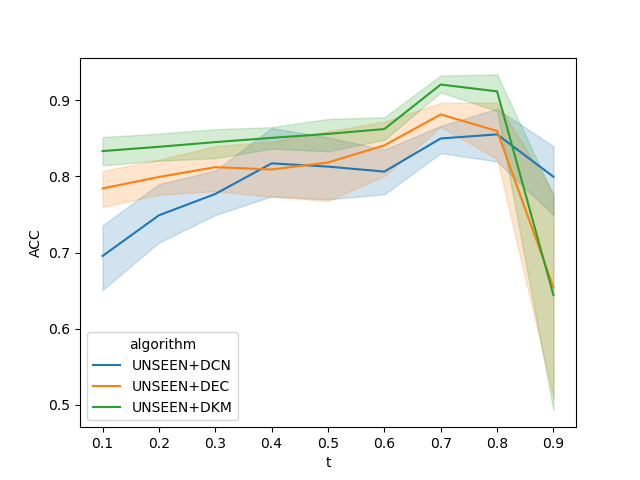}}
	\subfigure[USPS.]{\includegraphics[width=0.37\textwidth]{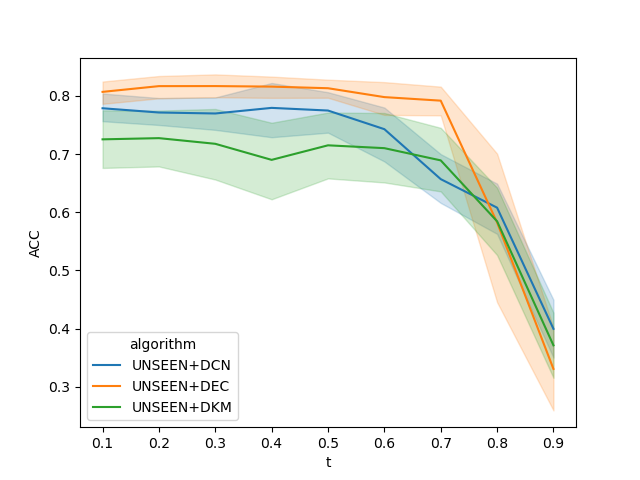}}
	\subfigure[MNIST.]{\includegraphics[width=0.37\textwidth]{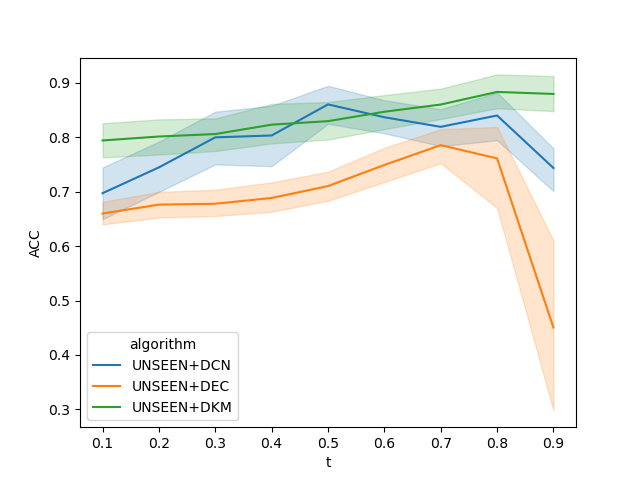}}
	\subfigure[Pendigits.]{\includegraphics[width=0.37\textwidth]{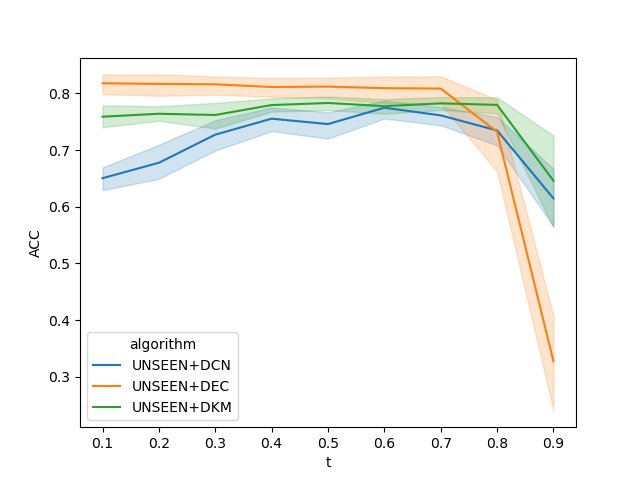}}
	\caption{The plots show clustering results regarding ACC for different values of the dying threshold $t$. The colored area marks the $95\%$ confidence interval.}
	\label{fig:dyingThreshold}
\end{figure*}

Last, we investigate the impact of the dying threshold $t$. To do this, we evaluate the performance of UNSEEN regarding ACC for values of $t \in [0.1, 0.9]$. We perform this experiment on the datasets Optdigits, USPS, MNIST, and Pendigits. 

When analyzing the plots shown in Figure~\ref{fig:dyingThreshold}, we see that most results are quite stable regarding the choice of $t$. Only UNSEEN+DCN shows strong fluctuations, where the ACC varies from $0.7$ to $0.86$ for MNIST, for example. Overall, performance appears to increase with increasing $t$. However, the value must not be set too high, as otherwise, too many clusters will be interpreted as dead, resulting in a crash in performance. 
From this, it can be concluded that $0.5$ represents a good trade-off between cluster quality and the risk of losing too many clusters. Hence, we chose this value for all other experiments.

\section{Conclusion}

To conclude, this paper presents UNSEEN, a framework for estimating the number of clusters that can be easily combined with many deep clustering algorithms. We demonstrate its performance using an extensive experimental evaluation that includes several image and tabular datasets. Furthermore, we conduct several ablations to justify the design choices and investigate the importance of UNSEEN's components. Several promising avenues for future work exist. For instance, we plan to examine how UNSEEN performs on unbalanced datasets. Another interesting direction is to test UNSEEN with a broader range of deep clustering algorithms, particularly those not based on centroids and based on contrastive learning.

\balance
\bibliographystyle{IEEEtranS}
\bibliography{references}

\begin{thebibliography}{10}
\providecommand{\url}[1]{#1}
\csname url@samestyle\endcsname
\providecommand{\newblock}{\relax}
\providecommand{\bibinfo}[2]{#2}
\providecommand{\BIBentrySTDinterwordspacing}{\spaceskip=0pt\relax}
\providecommand{\BIBentryALTinterwordstretchfactor}{4}
\providecommand{\BIBentryALTinterwordspacing}{\spaceskip=\fontdimen2\font plus
\BIBentryALTinterwordstretchfactor\fontdimen3\font minus \fontdimen4\font\relax}
\providecommand{\BIBforeignlanguage}[2]{{%
\expandafter\ifx\csname l@#1\endcsname\relax
\typeout{** WARNING: IEEEtranS.bst: No hyphenation pattern has been}%
\typeout{** loaded for the language `#1'. Using the pattern for}%
\typeout{** the default language instead.}%
\else
\language=\csname l@#1\endcsname
\fi
#2}}
\providecommand{\BIBdecl}{\relax}
\BIBdecl

\bibitem{pendigits}
E.~Alpaydin and F.~Alimoglu, ``{Pen-Based Recognition of Handwritten Digits},'' UCI Machine Learning Repository, 1996, {DOI}: https://doi.org/10.24432/C5MG6K.

\bibitem{optdigits}
E.~Alpaydin and C.~Kaynak, ``{Optical Recognition of Handwritten Digits},'' UCI Machine Learning Repository, 1998, {DOI}: https://doi.org/10.24432/C50P49.

\bibitem{andersonDarling}
T.~W. Anderson and D.~A. Darling, ``Asymptotic theory of certain" goodness of fit" criteria based on stochastic processes,'' \emph{The annals of mathematical statistics}, pp. 193--212, 1952.

\bibitem{autoencoder}
D.~H. Ballard, ``Modular learning in neural networks,'' in \emph{Proceedings of the 6th National Conference on Artificial Intelligence. Seattle, WA, USA, July 1987}.\hskip 1em plus 0.5em minus 0.4em\relax Morgan Kaufmann, 1987, pp. 279--284.

\bibitem{curseOfDimensionality}
R.~Bellman, ``Dynamic programming,'' \emph{Science}, vol. 153, no. 3731, pp. 34--37, 1966.

\bibitem{mdlBischof}
H.~Bischof, A.~Leonardis, and A.~Selb, ``{MDL} principle for robust vector quantisation,'' \emph{Pattern Anal. Appl.}, vol.~2, no.~1, pp. 59--72, 1999.

\bibitem{kmnist}
T.~Clanuwat, M.~Bober-Irizar, A.~Kitamoto, A.~Lamb, K.~Yamamoto, and D.~Ha, ``Deep learning for classical japanese literature,'' \emph{arXiv preprint arXiv:1812.01718}, 2018.

\bibitem{scde}
\BIBentryALTinterwordspacing
L.~Duan, C.~C. Aggarwal, S.~Ma, and S.~Sathe, ``Improving spectral clustering with deep embedding and cluster estimation,'' in \emph{2019 {IEEE} International Conference on Data Mining, {ICDM} 2019, Beijing, China, November 8-11, 2019}.\hskip 1em plus 0.5em minus 0.4em\relax {IEEE}, 2019, pp. 170--179. [Online]. Available: \url{https://doi.org/10.1109/ICDM.2019.00027}
\BIBentrySTDinterwordspacing

\bibitem{dbscan}
M.~Ester, H.~Kriegel, J.~Sander, and X.~Xu, ``A density-based algorithm for discovering clusters in large spatial databases with noise,'' in \emph{Proceedings of the Second International Conference on Knowledge Discovery and Data Mining (KDD-96), Portland, Oregon, {USA}}.\hskip 1em plus 0.5em minus 0.4em\relax {AAAI} Press, 1996, pp. 226--231.

\bibitem{dkm}
M.~M. Fard, T.~Thonet, and {\'{E}}.~Gaussier, ``Deep \emph{k}-means: Jointly clustering with \emph{k}-means and learning representations,'' \emph{Pattern Recognit. Lett.}, vol. 138, pp. 185--192, 2020.

\bibitem{pgmeans}
Y.~Feng and G.~Hamerly, ``Pg-means: learning the number of clusters in data,'' in \emph{Advances in Neural Information Processing Systems 19, Proceedings of the Twentieth Annual Conference on Neural Information Processing Systems, Vancouver, British Columbia, Canada, December 4-7, 2006}.\hskip 1em plus 0.5em minus 0.4em\relax {MIT} Press, 2006, pp. 393--400.

\bibitem{idec}
X.~Guo, L.~Gao, X.~Liu, and J.~Yin, ``Improved deep embedded clustering with local structure preservation,'' in \emph{Proceedings of the Twenty-Sixth International Joint Conference on Artificial Intelligence, {IJCAI} 2017, Melbourne, Australia, August 19-25, 2017}.\hskip 1em plus 0.5em minus 0.4em\relax ijcai.org, 2017, pp. 1753--1759.

\bibitem{dcec}
X.~Guo, X.~Liu, E.~Zhu, and J.~Yin, ``Deep clustering with convolutional autoencoders,'' in \emph{Neural Information Processing - 24th International Conference, {ICONIP} 2017, Guangzhou, China, November 14-18, 2017, Proceedings, Part {II}}, ser. Lecture Notes in Computer Science, vol. 10635.\hskip 1em plus 0.5em minus 0.4em\relax Springer, 2017, pp. 373--382.

\bibitem{gMeans}
G.~Hamerly and C.~Elkan, ``Learning the k in k-means,'' in \emph{Advances in Neural Information Processing Systems 16 [Neural Information Processing Systems, {NIPS} 2003, December 8-13, 2003, Vancouver and Whistler, British Columbia, Canada]}.\hskip 1em plus 0.5em minus 0.4em\relax {MIT} Press, 2003, pp. 281--288.

\bibitem{diptest}
J.~A. Hartigan and P.~M. Hartigan, ``The dip test of unimodality,'' \emph{The annals of Statistics}, pp. 70--84, 1985.

\bibitem{ari}
L.~Hubert and P.~Arabie, ``Comparing partitions,'' \emph{Journal of classification}, vol.~2, pp. 193--218, 1985.

\bibitem{usps}
J.~J. Hull, ``A database for handwritten text recognition research,'' \emph{IEEE Transactions on pattern analysis and machine intelligence}, vol.~16, no.~5, pp. 550--554, 1994.

\bibitem{dipmeans}
A.~Kalogeratos and A.~Likas, ``Dip-means: an incremental clustering method for estimating the number of clusters,'' in \emph{Advances in Neural Information Processing Systems 25: 26th Annual Conference on Neural Information Processing Systems 2012. Proceedings of a meeting held December 3-6, 2012, Lake Tahoe, Nevada, United States}, 2012, pp. 2402--2410.

\bibitem{adam}
\BIBentryALTinterwordspacing
D.~P. Kingma and J.~Ba, ``Adam: {A} method for stochastic optimization,'' in \emph{3rd International Conference on Learning Representations, {ICLR} 2015, San Diego, CA, USA, May 7-9, 2015, Conference Track Proceedings}, 2015. [Online]. Available: \url{http://arxiv.org/abs/1412.6980}
\BIBentrySTDinterwordspacing

\bibitem{nmi}
T.~O. Kvalseth, ``Entropy and correlation: Some comments,'' \emph{IEEE Transactions on Systems, Man, and Cybernetics}, vol.~17, no.~3, pp. 517--519, 1987.

\bibitem{mnist}
Y.~LeCun, L.~Bottou, Y.~Bengio, and P.~Haffner, ``Gradient-based learning applied to document recognition,'' \emph{Proceedings of the IEEE}, vol.~86, no.~11, pp. 2278--2324, 1998.

\bibitem{dipencoder}
C.~Leiber, L.~G.~M. Bauer, M.~Neumayr, C.~Plant, and C.~B{\"{o}}hm, ``The dipencoder: Enforcing multimodality in autoencoders,'' in \emph{{KDD} '22: The 28th {ACM} {SIGKDD} Conference on Knowledge Discovery and Data Mining, Washington, DC, USA, August 14 - 18, 2022}.\hskip 1em plus 0.5em minus 0.4em\relax {ACM}, 2022, pp. 846--856.

\bibitem{dipdeck}
C.~Leiber, L.~G.~M. Bauer, B.~Schelling, C.~B{\"{o}}hm, and C.~Plant, ``Dip-based deep embedded clustering with k-estimation,'' in \emph{{KDD} '21: The 27th {ACM} {SIGKDD} Conference on Knowledge Discovery and Data Mining, Virtual Event, Singapore, August 14-18, 2021}.\hskip 1em plus 0.5em minus 0.4em\relax {ACM}, 2021, pp. 903--913.

\bibitem{deepClusteringBenchmark}
C.~Leiber, L.~Miklautz, C.~Plant, and C.~B{\"{o}}hm, ``Benchmarking deep clustering algorithms with clustpy,'' in \emph{{IEEE} International Conference on Data Mining, {ICDM} 2023 - Workshops, Shanghai, China, December 4, 2023}.\hskip 1em plus 0.5em minus 0.4em\relax {IEEE}, 2023, pp. 625--632.

\bibitem{kmeansLloyd}
S.~P. Lloyd, ``Least squares quantization in {PCM},'' \emph{{IEEE} Trans. Inf. Theory}, vol.~28, no.~2, pp. 129--136, 1982.

\bibitem{deepect}
D.~Mautz, C.~Plant, and C.~B{\"{o}}hm, ``Deep embedded cluster tree,'' in \emph{2019 {IEEE} International Conference on Data Mining, {ICDM} 2019, Beijing, China, November 8-11, 2019}, J.~Wang, K.~Shim, and X.~Wu, Eds.\hskip 1em plus 0.5em minus 0.4em\relax {IEEE}, 2019, pp. 1258--1263.

\bibitem{acedec}
L.~Miklautz, L.~G.~M. Bauer, D.~Mautz, S.~Tschiatschek, C.~B{\"{o}}hm, and C.~Plant, ``Details (don't) matter: Isolating cluster information in deep embedded spaces,'' in \emph{Proceedings of the Thirtieth International Joint Conference on Artificial Intelligence, {IJCAI} 2021, Virtual Event / Montreal, Canada, 19-27 August 2021}.\hskip 1em plus 0.5em minus 0.4em\relax ijcai.org, 2021, pp. 2826--2832.

\bibitem{scikit-learn}
F.~Pedregosa, G.~Varoquaux, A.~Gramfort, V.~Michel, B.~Thirion, O.~Grisel, M.~Blondel, P.~Prettenhofer, R.~Weiss, V.~Dubourg, J.~Vanderplas, A.~Passos, D.~Cournapeau, M.~Brucher, M.~Perrot, and E.~Duchesnay, ``Scikit-learn: Machine learning in {P}ython,'' \emph{Journal of Machine Learning Research}, vol.~12, pp. 2825--2830, 2011.

\bibitem{xMeans}
D.~Pelleg and A.~W. Moore, ``X-means: Extending k-means with efficient estimation of the number of clusters,'' in \emph{Proceedings of the Seventeenth International Conference on Machine Learning {(ICML} 2000), Stanford University, Stanford, CA, USA, June 29 - July 2, 2000}.\hskip 1em plus 0.5em minus 0.4em\relax Morgan Kaufmann, 2000, pp. 727--734.

\bibitem{ddc}
Y.~Ren, N.~Wang, M.~Li, and Z.~Xu, ``Deep density-based image clustering,'' \emph{Knowl. Based Syst.}, vol. 197, p. 105841, 2020.

\bibitem{deepdpm}
M.~Ronen, S.~E. Finder, and O.~Freifeld, ``Deepdpm: Deep clustering with an unknown number of clusters,'' in \emph{{IEEE/CVF} Conference on Computer Vision and Pattern Recognition, {CVPR} 2022, New Orleans, LA, USA, June 18-24, 2022}.\hskip 1em plus 0.5em minus 0.4em\relax {IEEE}, 2022, pp. 9851--9860.

\bibitem{silhouetteScore}
P.~J. Rousseeuw, ``Silhouettes: a graphical aid to the interpretation and validation of cluster analysis,'' \emph{Journal of computational and applied mathematics}, vol.~20, pp. 53--65, 1987.

\bibitem{stopElbow}
E.~Schubert, ``Stop using the elbow criterion for k-means and how to choose the number of clusters instead,'' \emph{{SIGKDD} Explor.}, vol.~25, no.~1, pp. 36--42, 2023.

\bibitem{aec}
C.~Song, F.~Liu, Y.~Huang, L.~Wang, and T.~Tan, ``Auto-encoder based data clustering,'' in \emph{Progress in Pattern Recognition, Image Analysis, Computer Vision, and Applications - 18th Iberoamerican Congress, {CIARP} 2013, Havana, Cuba, November 20-23, 2013, Proceedings, Part {I}}, ser. Lecture Notes in Computer Science, vol. 8258.\hskip 1em plus 0.5em minus 0.4em\relax Springer, 2013, pp. 117--124.

\bibitem{gapStatistic}
R.~Tibshirani, G.~Walther, and T.~Hastie, ``Estimating the number of clusters in a data set via the gap statistic,'' \emph{Journal of the Royal Statistical Society: Series B (Statistical Methodology)}, vol.~63, no.~2, pp. 411--423, 2001.

\bibitem{tsne}
L.~Van~der Maaten and G.~Hinton, ``Visualizing data using t-sne.'' \emph{Journal of machine learning research}, vol.~9, no.~11, 2008.

\bibitem{fmnist}
H.~Xiao, K.~Rasul, and R.~Vollgraf, ``Fashion-mnist: a novel image dataset for benchmarking machine learning algorithms,'' \emph{arXiv preprint arXiv:1708.07747}, 2017.

\bibitem{dec}
J.~Xie, R.~B. Girshick, and A.~Farhadi, ``Unsupervised deep embedding for clustering analysis,'' in \emph{Proceedings of the 33nd International Conference on Machine Learning, {ICML} 2016, New York City, NY, USA, June 19-24, 2016}, ser. {JMLR} Workshop and Conference Proceedings, vol.~48.\hskip 1em plus 0.5em minus 0.4em\relax JMLR.org, 2016, pp. 478--487.

\bibitem{clusteringSurvey}
D.~Xu and Y.~Tian, ``A comprehensive survey of clustering algorithms,'' \emph{Annals of Data Science}, vol.~2, pp. 165--193, 2015.

\bibitem{dcn}
B.~Yang, X.~Fu, N.~D. Sidiropoulos, and M.~Hong, ``Towards k-means-friendly spaces: Simultaneous deep learning and clustering,'' in \emph{Proceedings of the 34th International Conference on Machine Learning, {ICML} 2017, Sydney, NSW, Australia, 6-11 August 2017}, ser. Proceedings of Machine Learning Research, vol.~70.\hskip 1em plus 0.5em minus 0.4em\relax {PMLR}, 2017, pp. 3861--3870.

\bibitem{acc}
Y.~Yang, D.~Xu, F.~Nie, S.~Yan, and Y.~Zhuang, ``Image clustering using local discriminant models and global integration,'' \emph{IEEE Transactions on Image Processing}, vol.~19, no.~10, pp. 2761--2773, 2010.

\bibitem{deepClusteringSurvey}
S.~Zhou, H.~Xu, Z.~Zheng, J.~Chen, Z.~Li, J.~Bu, J.~Wu, X.~Wang, W.~Zhu, and M.~Ester, ``A comprehensive survey on deep clustering: Taxonomy, challenges, and future directions,'' \emph{CoRR}, vol. abs/2206.07579, 2022.

\end{thebibliography}

\end{document}